%% file: skeletonnet.tex
\documentclass[runningheads]{llncs}

\usepackage{graphicx}
\usepackage{amsmath,amssymb} % define this before the line numbering.

\usepackage{color}
\usepackage{hyperref}       % hyperlinks
\usepackage{url}   
\usepackage{tabularx}
\usepackage{comment}

\begin{document}

\title{Skeleton Transformer Networks: 3D Human Pose and Skinned Mesh from Single RGB Image}

\titlerunning{Skeleton Transformer Networks}
\author{Yusuke Yoshiyasu, Ryusuke Sagawa, Ko Ayusawa, Akihiko Murai}

%\orcidID{2222--3333-4444-5555}
%Please include author names in full in the paper, 
%If any authors have names that can be parsed into FirstName LastName in multiple ways, please include the correct parsing, in a comment to the volume editors:
%\index{Lastnames, Firstnames}

\authorrunning{Y. Yoshiyasu et al.} % A shorter version of authors' name

%===========================================================

\institute{National Institute of Advanced Industrial Science and Technology (AIST) \and CNRS-AIST JRL (Joint Robotics Laboratory), UMI3218/RL,	\email{\{yusuke-yoshiyasu, ryusuke.sagawa, k.ayusawa, a.murai\}@aist.go.jp}} 
%\def\ACCV18SubNumber{483}
%\author{Anonymous ACCV 2018 submission}
%\institute{Paper ID \ACCV18SubNumber}
%\begin{comment}

\maketitle

%===========================================================
\begin{abstract}
	In this paper, we present Skeleton Transformer Networks (SkeletonNet), an end-to-end framework that can predict not only 3D joint positions but also 3D angular pose (bone rotations) of a human skeleton from a single color image. This in turn allows us to generate skinned mesh animations. Here, we propose a two-step regression approach. The first step regresses bone rotations in order to obtain an initial solution by considering skeleton structure. The second step performs refinement based on heatmap regressor using a 3D pose representation called cross heatmap which stacks heatmaps of xy and zy coordinates. By training the network using the proposed 3D human pose dataset that is comprised of images annotated with 3D skeletal angular poses, we showed that SkeletonNet can predict a full 3D human pose (joint positions and bone rotations) from a single image in-the-wild. 
	%Furthermore, by exploiting skeletal structure, SkeletonNet can obtain reasonably accurate results in a less data situation where 3D human pose dataset is available, which contains a small number of human subjects. 
\end{abstract}

%===========================================================
%===========================================================
\input{intro}

\input{related}

\input{method}
\input{result}

\input{conclusion}

\begin{figure}[h]
	\centering
	\includegraphics[width=1\linewidth]{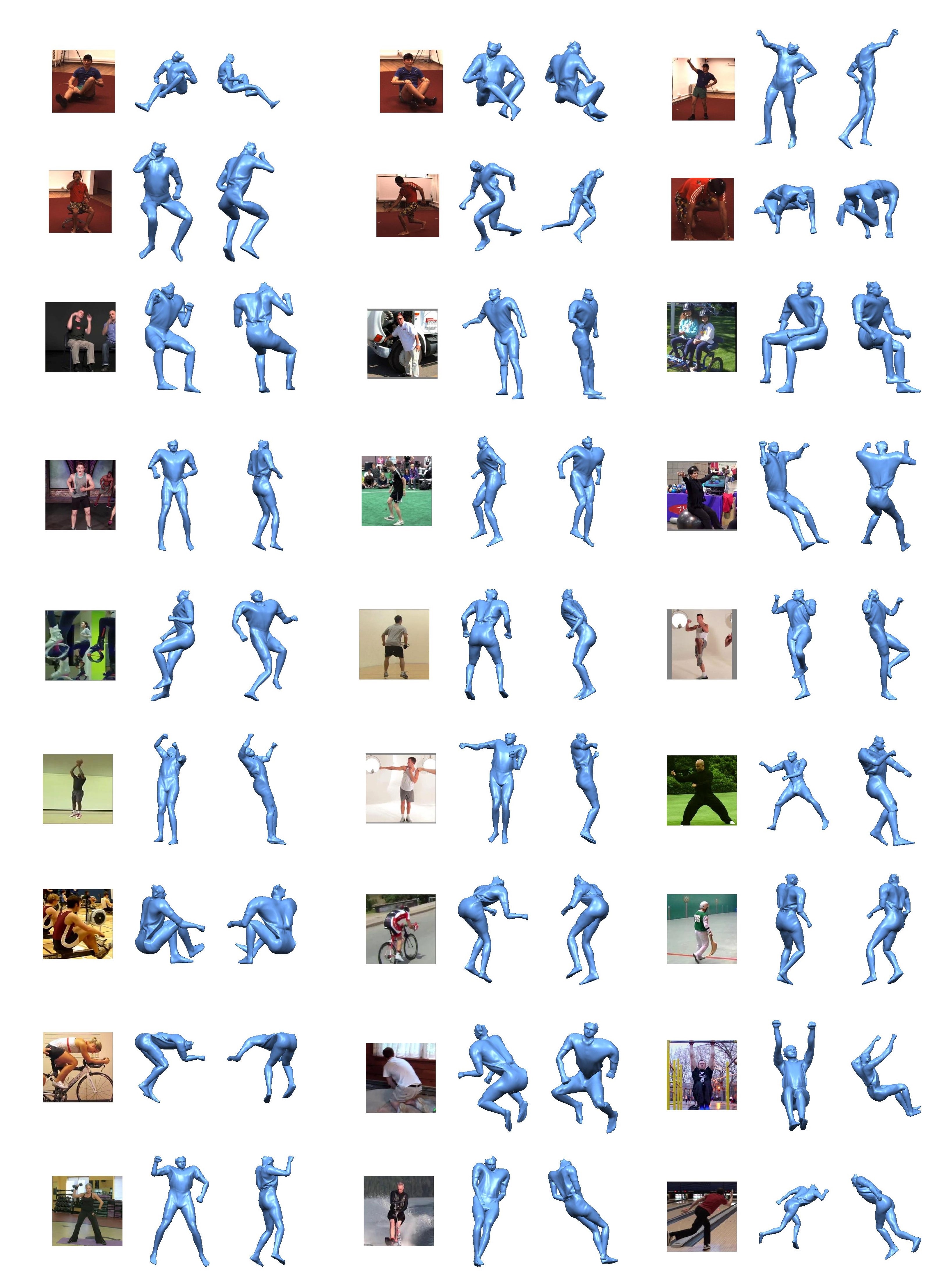}
	\caption{More results.}
	\label{fig:result_mesh2}
\end{figure}

%\bibliographystyle{ieee}
%\bibliography{egbib}

\section*{Acknowledgment}
I would like to thank Rie Nishihama and CNRS-AIST JRL members for supporting us constructing 3D human pose dataset. This work was partly supported by JSPS Kakenhi No. 17K18420 and No. 18H03315.     

%this would normally be the end of your paper, but you may also have an appendix
%within the given limit of number of pages
%\end{comment}
%\input{result}
\bibliographystyle{ieee}
\bibliography{egbib}
\end{document}

%% file: intro.tex
\section{Introduction}
Estimating 3D human pose from a single image is an important yet very challenging problem in computer vision, where applications range  from surveillance to robotics. Recent work has shown that convolutional neural networks (ConvNets) can detect 2D joint positions accurately. The key to achieving accurate predictions is to represent 2D joint locations as heatmaps and iteratively refines them gradually by incorporating context information \cite{newell2016stacked,belagiannis2016recurrent}--early approaches  \cite{toshev2014deeppose} on the other hand directly perform regression of the 2D joint coordinates using ConvNets, which is a difficult problem that needs to model a highly-nonlinear mapping from an input image to real values. Recent techniques, such as OpenPose \cite{cao2017realtime}, can robustly detect 2D joints of multiple people in an image.  

%Early approaches directly regresses the 2D joint coordinates using ConvNets \cite{toshev2014deeppose}. Since regressing the 2D joint locations from an image is a difficult problem that is highly nonlinear, 

In contrast to its 2D counterpart, the progress of 3D human pose detection has been relatively slow. The main challenges regarding to 3D human pose estimation is as follows: 

\noindent {\bf 3D pose representation } To predict 3D joint locations using ConvNets, 3D pose representation used is critical, which affects prediction accuracy. Previous approaches have shown that regression of a 3D pose using heatmaps (e.g., volumetric \cite{PavlakosZDD16} and 2D heatmaps + depth \cite{zhou2017weakly}) leads to accurate 3D joint predictions. On the other hand, regression of joint angles using ConvNets \cite{zhou2016deep,hmrKanazawa17} has not been successful so far in contributing to accurate 3D joint localizations, because they are difficult to learn with ConvNets due to their high non-linearity. From the application point of view, such as computer animation and biomechanics, it is desirable to predict not only 3D joint locations but also angular pose of skeleton, e.g., joint angles or segment rotations.     

\noindent {\bf Data scarcity } Compare to 2D human pose dataset, 3D human pose dataset is smaller in size. This is because obtaining a 3D human pose dataset where the images paired with 3D joint annotations requires time and effort. In particular, annotations of 3D angular skeletal poses are difficult to obtain. One common way to achieve this is to use a motion capture (MoCap) system and RGB video cameras at the same time. However, such dataset are usually limited to a small variety of subjects---for example, Human 3.6M dataset \cite{ionescu2014human3}, which is the most common dataset for 3D human pose, is limited to around 10 subjects. Consequently, it is difficult to learn sufficient visual features from 3D pose dataset solely to localize 2D/3D joints accurately. 

% how to cope with data scarsity

%In terms of joint representation, previous approaches can be roughly divided into three groups: The first group represent joints as 3D points. It directly regress 3D joint locations from ConvNet features. However, this approach is very difficult as the mapping from an image to 3D joint positions is highly nonlinear. The second approach first predicts 2D joint positions and then regress 3D joint positions or depths from them. The third approach uses a volumetric heatmap representation, which is an extension of its 2D counter part. As this volumetric approach can avoid regressing the real values in a highly nonlinear manner, it is a natural representation for 3D joints and can accurately predict them. %However, this approach requires a large amount of parameters and memory spaces to store feature maps as it additional dimension.   
	
% structure 
Skeletal structure has been incorporated into 3D human pose estimation as a form of constraints or subspace. In biomechanics and robotics, forward and inverse kinematics have been well-studied and are used to generate human pose from MoCap by controlling joint angles of a skeleton. Previous approaches in the computer vision field estimated 3D human pose from 2D key points using a statistical model and enforcing constraints such as segment length \cite{simo2012single}, joint limit \cite{akhter2015pose} and symmetry. In computer graphics, human skeletal pose is often represented using linear or affine transformation matrices, as can be seen for example in linear blend skinning for character animation.  % and other deformable models.

%On the other hand, ConvNets incorporates structure via learning of image features. This can be done indirectly by for example  incorporating pairwise potentials using a graphical model or recurrence which makes predictions gradually better. However, it is known that ConvNets are translation invariant and because of this property ConvNets does not explicitly respect spatial structure in the image. In the case of human pose estimation, this leads to, for example, confusions in the left hand and right hand, which is especially true under the condition of a limited amount of image variations like 3D human pose dataset. Some methods perform regression to predict joint angles but the high nonlinearity prevents us from accurate prediction of joint locations.  

In this paper, we propose skeleton transformer networks (SkeletonNet) for 3D human pose detection which respects skeletal structure while attaining 3D joint prediction accuracy. SkeletonNet combines and benefits from the two paradigms, skeleton and heatmap representations. SkeletonNet first regresses bone rotations to an input image in order to have an initial solution which is not precisely accurate but considers skeletal structure. Starting from this initial solution, the second step refines it using ConvNet heamap regressor. This strategy allows us to recover reasonably accurate predictions of full 3D human poses (joint positions and bone rotations) from a single in-the-wild image. To contribute to solving the data scarcity problem, we also construct a dataset where 3D angular skeletal  poses are annotated on in-the-wild images, based on a human-validation approach. By fusing the proposed dataset and Mocap dataset captured under a controlled environment (Human3.6M), SkeletonNet can predict a full 3D pose (joint positions and bone rotation) from a single image in-the-wild. In addition, experimental results showed that SkeletonNet outperforms previous approaches based on joint angles \cite{hmrKanazawa17,zhou2016deep} in terms of MPJPE joint position accuracy.  

%In a less data situation where e.g., training dataset that contains several subjects is only available, our method outperforms previous approaches by leveraging skeletal structure. Furthermore, our approach can predict angular pose, which is useful for applications like computer animation and biomechanics.

%The first step regresses angular pose in the form of segment rotation matrices, which is a linear representation that is relatively easier to regress with ConvNets.

The contributions of this work is summarized as follows:

\begin{itemize}
\item 
We propose an end-to-end ConvNet framework for predicting a full 3D human pose (joint positions and bone rotations). 		
	
\item 
We propose a bone rotation regressor which predicts 3D human pose using $3\times3$ transformation matrices. To make arbitrarily linear transformations into rotations, we propose a Gram Schmidt orthogonalization layer. This  combination is the key to learning angular pose accurately with using ConvNets.

\item 
We propose a 3D human pose representation called cross heatmap for accurate 3D joint detection. This representation combines two heat maps, one for representing 2D joints in image space (xy space) and the other for zy space. The benefit of this representation is that it is more efficient than the volumetric heatmaps \cite{PavlakosZDD16}, while accurately predicting 3D joint positions when trained using Mocap-video dataset (Human 3.6M) and in-the wild dataset (MPII with 3D pose annotations).  

\item 
We built a 3D human pose dataset in-the-wild which includes annotations of 3D bone rotations.     
 
%\item 
%We also build a 3D human pose dataset that includes 3D joint positions and rotations, which is constructed from in-the wild images by providing 3D annotations based on a human-in-the-loop strategy.  
\end{itemize}
  
%In this paper, we propose a 3D human pose estimation technique that can accurately estimate 3D joint position efficiently. The key to the accurate detection of human pose is the method that iteratively refines 3D joint angles. We call this architecture as kinematic layers. To exploit image features in 3D pose estimation, we first obtain 3D joint annotations (joint positions and angles) from 2D joint annotations. The network is trained based on human pose image datasets with coupled with these 3D annotations. 

%*lifting from deep  
%*coarse-to-fine volumetric
%*weak supervision

%% file: related.tex
\section{Related work}
\noindent {\bf Estimating 2D joint positions using ConvNets } Recent work has shown that the detection of 2D joint positions can be done very accurately using convolutional neural networks (ConvNets). Toshev et al. \cite{toshev2014deeppose} first proposed a method based on ConvNets for detecting human pose i.e., 2D key points representing joint locations from a single image.  Tompson et al. instead represented joint locations in images using 2D heat maps so that it can avoid complicated a nonlinear mapping that goes from an image to xy pixel coordinates. The recent techniques, such as the stacked hourglass network \cite{newell2016stacked} and its variants \cite{DBLP:journals/corr/abs-1708-01101,DBLP:journals/corr/ChuYOMYW17,DBLP:journals/corr/ChenSWLY17}, accurately predict 2D joint positions by iteratively refining 2D heatmaps.

\noindent {\bf Predicting 3D joint positions}
The early approaches predicts 3D joint positions from key points \cite{ramakrishna2012reconstructing}. These approaches assume that the almost perfect 2D key points are already extracted from an image. Li et al. \cite{li20143d} first used ConvNets to directly regress 3D human joints with an image. There are two main reasons for the improvements on accuracy of 3D human pose detection. First, the recent approach combines multiple data sources to increase the 3D pose dataset \cite{mehta2016monocular,zhou2017weakly}. Second, the recent techniques make use of more natural 3D joint representation. For example, Pavlakos et al. \cite{PavlakosZDD16} uses a volumetric heatmap representation, which can avoid regressing the real values in a highly nonlinear manner. Other methods first predicts 2D joints with heatmaps and then regress 3D joint positions or depths from them. Tome et al. \cite{tome2017lifting} have tried to iteratively update 3D joints represented as a weighted combination of PCA basis that is constructed from 3D MoCap dataset. 

\noindent {\bf Exploiting skeletal structure and predicting angular pose }
In biomechanics, robotics and computer animation fields, inverse kinematics has been well-studied and used to generate human pose from MoCap by controlling joint angles. Previous approaches \cite{ramakrishna2012reconstructing,bogo2016keep} estimated 3D human pose from 2D key points by combining a statistical model and constraints such as joint limit \cite{akhter2015pose}, segment length \cite{ramakrishna2012reconstructing} and symmetry. Some methods perform regression of joint angles or axis angles \cite{zhou2016deep,hmrKanazawa17} to estimate angular skeletal pose using ConvNets but the high nonlinearity prevents them from accurate prediction of joint locations.

\noindent {\bf Weakly supervision and predictions from in-the-wild images } 
Recent works tackle the data scarcity problem by using both 3D human pose dataset captured in the experimental room and 2D human pose dataset captured in a wide range of environment \cite{zhou2017weakly,mehta2016monocular,sun2017compositional}. Sun et al. \cite{sun2017compositional} used compositional loss function that is defined by integrating 2D positions and depths by properly normalizing the regression target values. Zhou et al. \cite{zhou2017weakly} took a weak supervised approach and used bone length constraint when 3D information (depths) is not available. The use of this approach not only enables 3D joint predictions from in-the wild images but also performance boost in joint prediction accuracy. We go further by building a 3D human pose dataset in-the-wild which includes annotations of 3D bone rotations.

%% file: method.tex
\begin{figure}[tb]
	\centering
	\includegraphics[width=1\linewidth]{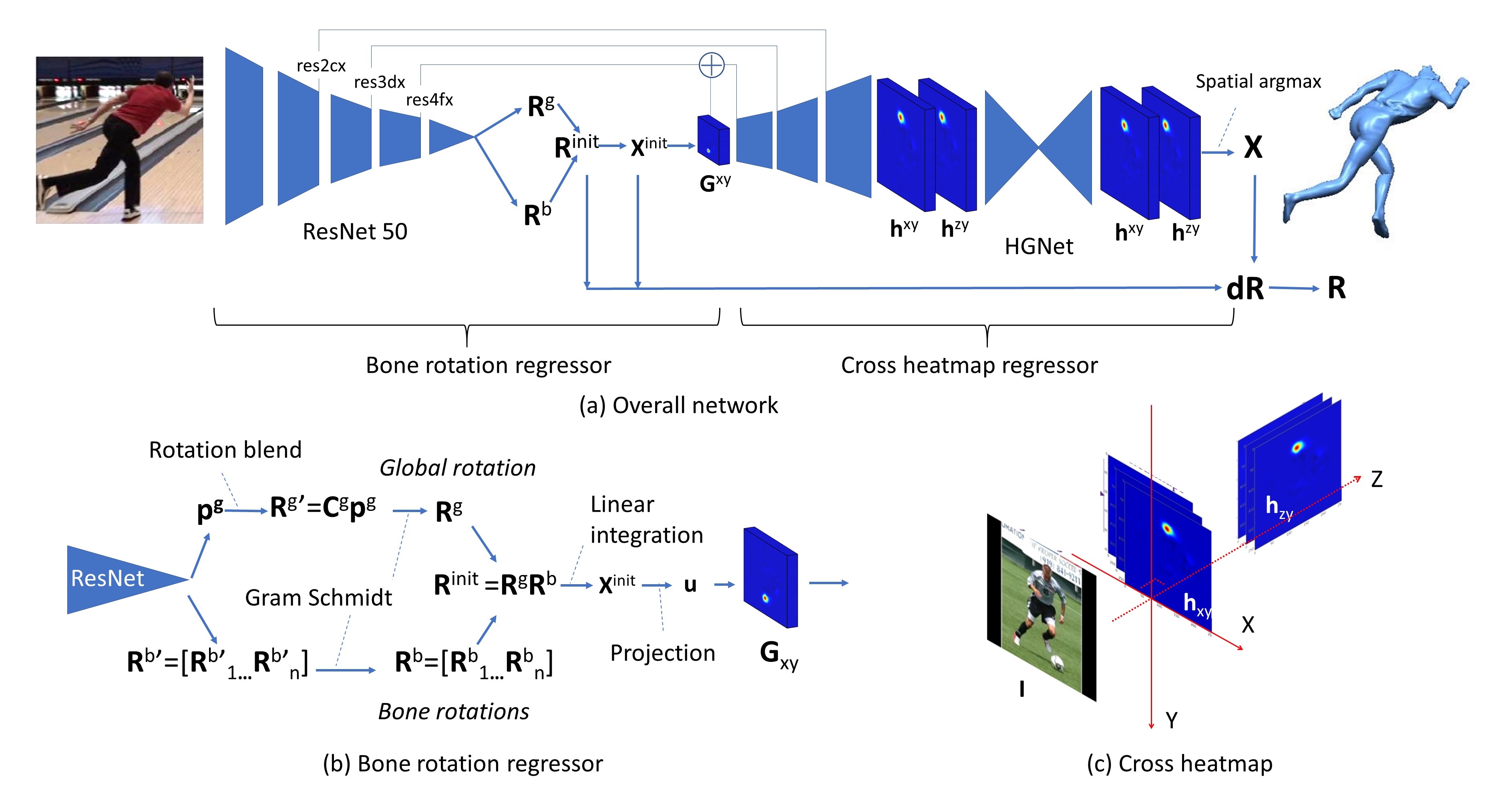}
	\caption{Network architecture of SkeletonNet. (a) SkeletonNet is a two-step regressor. (b) The first part performs regression of bone rotations. (c) The second part detects heatmaps of $xy$ and $zy$ spaces. }
	\label{fig:net}
\end{figure}

\section{Skeleton Transformer Networks}

Skeleton Transformer Networks (SkeletonNet) is based on deep ConvNets, which predicts not only 3D joint positions but also rotations of body segments. Our approach is based on an end-to-end two-step regression approach. The network architecture of SkeletonNet is depicted in Fig. \ref{fig:net} (a). The first step is called {\it bone rotation regressor} that performs regression of  angular skeletal pose using $3 \times 3$ rotation matrices to provide an initial solution. The resulting pose is not precisely accurate but it respects skeletal structure, without having to produce large errors such as left and right confusions. The second step, {\it cross heatmap regressor}, starts from this initial solution and refine the joint predictions using heatmap regression. We benefit from the two different paradigms, i.e., skeleton and ConvNets, to achieve accuracy and preservation of structure. % In this way, visual features are aggregated from global to local scales, leading to reasonably accurate 3D human pose detection.

\subsection{Bone rotation regressor}
\label{sec:skeletal}
Bone rotation regressor predicts bone rotations of a human skeleton. We achieve this by solving two simpler problems separately: predicting 1) a global rotation and 2) transformations of bone segments relative to the root. The idea behind this strategy is that the global orientation of the body has some discrete patterns e.g., sit, stand and lie, which can be effectively solved as a classification problem. On the other hand, bone rotations have more continuous distributions within some range, which can be effectively modeled as a regression problem.    

To predict a global rotation, we convert the rotation estimation problem into a classification problem. Specifically, we first cluster the training dataset into 200 clusters based on its global rotations with $k$-means clustering. The samples within the same cluster is put in the same class. ConvNets is trained to output rotation class probabilities ${\bf p}^{\rm g}$ using Softmax. We use a classification loss (cross entropy) for supervision:
\begin{eqnarray}
{\cal L}_{\rm RotG} = L_{cls}({\bf p}^{\rm g}, \bar{{\bf p}}^{\rm g}) \nonumber
\end{eqnarray}
where $L_{cls}$ is the log loss and $\bar{{\bf p}}^{\rm g}$ is the one-hot class label of global rotation. From the output probabilities ${\bf p}^{\rm g}$, we obtain a $3\times3$ global transformation matrix by blending cluster centers ${\bf C}^{\rm g}$, ${\bf R}^{\rm g'} = {\bf C}^{\rm g} {\bf p}^{\rm g}$. We tried linear blending of axis angles but found that blending matrices works better.

Since bone rotations relative to the root have more continuous distributions than global rotation, regression is more suitable in this case than classification. Bone rotations ${\bf R}^{\rm b'} = [{\bf R}^{\rm b'}_1 \ldots {\bf R}^{\rm b'}_n]$, where $n$ is the number of bones, are thus predicted directly using a $3 \times 3$ rotation matrix (9 parameters). The loss for bone rotations is defined using the mean squared error (MSE) loss as: 
\begin{eqnarray}
{\cal L}_{\rm RotB} = \sum_i^n ||{\rm vec}({\bf R}^{\rm b'}_i) - {\rm vec}(\bar{{\bf R}}^{\rm b}_i)||_2^2 \nonumber
\end{eqnarray}
where ${\rm vec}(\cdot)$ makes a matrix to a vector and $\bar{{\bf R}}^{\rm b}$ is the ground truth bone rotations. Note that regression of rotation matrices demands more memory spaces because they need more parameters than other rotation representation, such as Euler angles, quaternions and axis angles. However, for 3D human pose detection, we have under 20 joints to predict, which means that the additional costs are almost negligible. The down side of Euler angles and quaternions are their nonlinearities and ambiguities (sign flips for quaternions and periodical angle jumps for Euler angles), which is difficult to use as the regression targets---we could not train a network properly using Euler angles as supervisions as reported in \cite{zhou2016deep}.

% joint rotation constraints
\subsubsection{Gram Schmidt orthogonalization layer}
The problem of the above strategy is that it does not guarantee to produce orthonormal matrices. This means the resulting skeleton is deformed in an undesirable way with scales and shears. To solve this issue, we propose the Gram Schmidt (GS) orthogonalization layer which performs GS to make transformations into rotations. GS requires elemental functions only, such as dot product, subtraction and division, which is differentiable and can be relatively easily incorporated into ConvNets. We input global transformation ${\bf R}^{\rm g'}$ and bone transformations ${\bf R}^{\rm b'}$ into the GS layer to make transformations to rotations, obtaining ${\bf R}^{\rm g}$ and ${\bf R}^{\rm b}$. 

Once transformations are orthonormalized, we multiply a global rotation ${\bf R}^{\rm g}$ and bone rotations ${\bf R}^{\rm b}$ in order to obtain the absolute bone rotations. Finally, 3D joint positions are computed by applying these absolute rotations to the original bone vectors in the rest pose and performing linear integration to add up bone vectors from the root (Fig. \ref{fig:net} (b)).% This step can be done with a single step by: $ $. 
      
\subsection{Cross heatmap regressor}
We propose cross heatmap regressor which is used for refining the 3D joints obtained using bone rotation regressor. In the current design, cross heatmap regressor stacks xy and zy heatmaps (Fig. \ref{fig:net} (c)) because they sufficiently cover xyz coordinates and the variance of human joint locations in zy coordinates are usually larger than that of zx space.

To integrate bone rotation regressor and cross heatmap regresor, we project 3D joint positions obtained in Sec \ref{sec:skeletal} into the image plane (xy plane). Here, we did not estimate a camera pose and scale explicitly. Instead we scale the xy coordinates to 90\% of the width of the first upsampling layer, which is 16 pixels. From the projected 2D joints,  2D Gaussian maps \cite{PavlakosZDD16} are obtained and, after convolutions, they are summed up with the feature maps from bone rotation regressor to serve as approximate positions of 2D joints for cross heatmap regressor. Note that all of these process are differentiable, which can be optimized using back propagation. 

Once the feature maps are up-sampled three times to the size of $64\times64$, a single stack of hourglass module \cite{newell2016stacked} is used to compute cross heatmaps. The cross heatmap representation concatenates two heatmaps, one for representing 2D joints in image space (xy space) ${\bf h}^{\rm xy}$ and the other for the zy space, ${\bf h}^{\rm zy}$. For training we use the MSE loss as:
\begin{eqnarray}
{\cal L}_{\rm hm} = \sum_i^m \sum_{j,k} ||{\bf h}^{\rm xy}_{(j,k)} - \bar{{\bf h}}^{\rm xy}_{(j,k)} ||_2^2 + \sum_i^m \sum_{j,k} ||{\bf h}^{\rm zy}_{(j,k)} - \bar{{\bf h}}^{\rm zy}_{(j,k)} ||_2^2 \nonumber
\end{eqnarray}
where $\bar{{\bf h}}^{\rm xy}$ and $\bar{{\bf h}}^{\rm zy}$ are the ground truth heatmaps for xy and zy spaces. In addition, $m$ is the number of joints. The benefit of this representation is that it is more compact and efficient than the volumetric heatmaps, while maintaining accuracy. To extract xyz coordinates from cross heatmaps in an end-to-end manner, we use spatial argmax layers similar to those proposed in  \cite{luvizon2017spatial_argmax,levine2015visuo_motor}. Finally, we compute rotations ${\bf dR}$ that align bone vectors obtained using bone rotation regressor with that of the final predicted positions. They are multiplied with the predicted absolute bone rotations ${\bf R}^{\rm init}$ to make them consistent with the final joint positions ${\bf x}$. This way, we can exploit accurate heatmaps to refine rotations, avoiding a difficult regression problem of nonlinear angle parameters. Now that the loss for the final positions ${\bf x}$ and rotations ${\bf R}$ are defined as: 
\begin{eqnarray}
{\cal L}_{\rm pos} = \sum_i^m ||{\bf x}_i - \bar{{\bf x}}_i||_2^2, \; \;
{\cal L}_{\rm Rot} = \sum_i^n ||{\rm vec}({\bf R}_i) - {\rm vec}(\bar{{\bf R}}_i)||_2^2  \nonumber
\end{eqnarray}
where and $\bar{{\bf x}}_i$ and $\bar{{\bf R}}_i$ are the ground truth labels of positions and rotations. 

With the final positions ${\bf x}$ and rotations ${\bf R}$, linear blend skinning can be done to produce a 3D mesh. Note that we perform skinning outside the network but this process can also be done within the network in an end-to-end manner, as this is a linear matrix multiplication. 

\subsection{Loss function} For supervision, a standard cross entropy loss and MSE loss is used for comparing the predictions and ground truth labels of the global rotation class probability, bone rotations and cross heatmaps. In total, we minimize the loss function of the form:
\begin{equation}
{\cal L}_{\rm total} =  {\cal L}_{\rm RotG} + \alpha {\cal L}_{\rm RotB} + \beta {\cal L}_{\rm Rot}  +  \gamma {\cal L}_{\rm pos} + \lambda {\cal L}_{\rm hm} \nonumber
\end{equation}
where ${\cal L}_{\rm RotG}$, ${\cal L}_{\rm RotB}$, ${\cal L}_{\rm Rot}$, ${\cal L}_{\rm pos}$ and ${\cal L}_{\rm hm}$ are a cross entropy loss for global rotation and MSE loss for bone rotations ${\bf R}^{\rm b}$, final rotations ${\bf R}$, final positions ${\bf x}$ and cross heatmaps (${\bf h}^{\rm xy}$ and ${\bf h}^{\rm zy}$), respectively. In addition, $\alpha$, $\beta$, $ \gamma$  and  $\lambda$ are the respective weights.  
%\begin{figure}[tb]
%	\centering
%	%\includegraphics[width=1\linewidth]{image/technical%_component.jpg}
%	\caption{Network components. }
%	\label{fig:crossHM}
%\end{figure}

\begin{figure}[tb]
	\centering
	\includegraphics[width=1\linewidth]{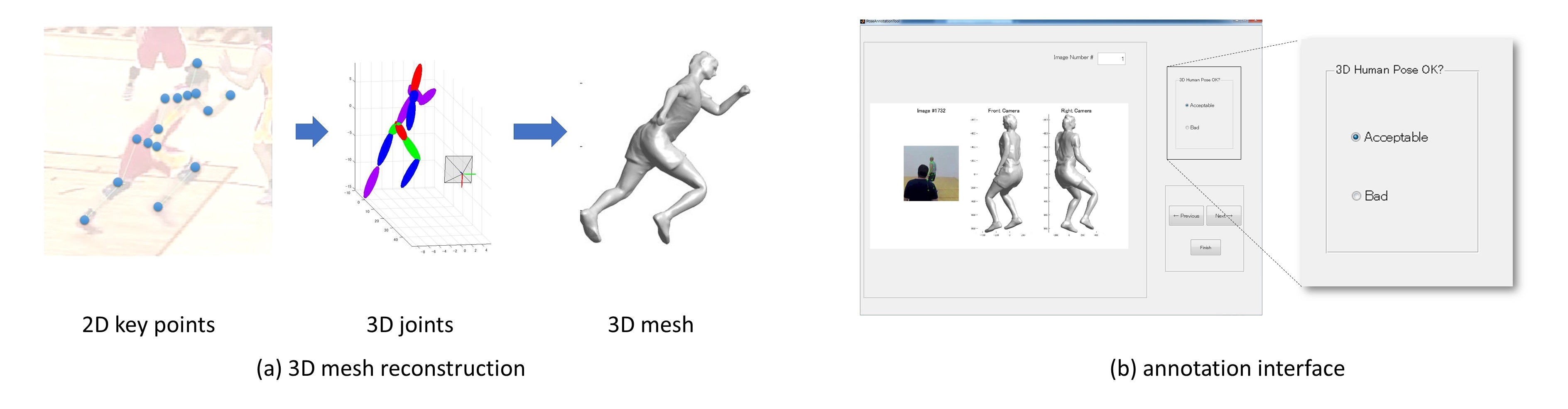}

	\caption{3D pose annotation system. (a) Given 2D key points as inputs, we compute 3D joints using the projected matching pursuit (PMP) technique. Next, skeleton rotations are obtained using a skeleton transformation optimization method. 3D mesh is obtained from rotations and positions. (b) With the annotation tool, the annotator is just needed to judge whether the 3D pose is acceptable or not by comparing the real image and the rendered image created from the mesh. }
	\label{fig:annotation_tool}
\end{figure}

\section{In-the-wild 3D Human Pose Dataset }
\label{sec:dataset}

To build 3D human pose dataset that have variations in clothes and background, we annotate 3D positions and bone rotations on MPII human pose dataset as shown in Fig. \ref{fig:annotation_tool}. To this end, we followed the human verification approach \cite{DBLP:journals/corr/PapadopoulosUKF16}. Here a 3D human pose is obtained from 2D key points using a previous technique, which will then be sorted by human annotators whether each result is acceptable or not. This is in spirit similar to the concurrent work of \cite{Lassner:UP:2017}. The aim for constructing our dataset is not to provide accurate 3D pose annotations but to obtain reasonable ones on in-the-wild images, including joint positions and rotations, such that they can be used for training our bone rotation regressor. In fact, the annotators are instructed to judge whether the pose is `acceptable' or `bad', e.g., if the global rotation of the resulting pose looks deviating from the true pose more than 30 deg, the pose is `bad'. As a consequence, our dataset have more 3D pose annotations than \cite{Lassner:UP:2017}, possibly at the cost of accuracy.    

To obtain 3D joint positions from 2D key points, we use the projected matching pursuit (PMP) approach \cite{ramakrishna2012reconstructing}. This method calculates a camera pose, scale and 3D joint positions as a combination of PCA basis that is constructed from Mocap database. From the resulting 3D joint positions, rotations of bones are obtained based on a method which is conceptually similar non-rigid surface deformation techniques \cite{Sorkine:2007:ASM:1281991.1282006}. Specifically, the skeleton in the rest shape is fitted to the PMP result by balancing the rigidity of bones, the smoothness between bone rotations and the position constraints to attract the skeleton to them. The initial rotations are obtained from local coordinate frames, which are defined in a similar manner using \cite{akhter2015pose}. Reconstructing one model from 2D key points takes approx. 1min.     

We also designed a simple annotation tool Fig. \ref{fig:annotation_tool} (b) to simplify the process of 3D pose annotations. With this tool, human annotators are just required to decide a 3D pose is acceptable or not. In addition, we obtained skin meshes from the 3D pose using linear blend skinning and showed rendered images. By visualizing a skin mesh, it makes the annotators' decisions significantly easier and quicker. From among the images in MPII dataset, we extracted those with all 16 joints are inside the image region, which was approximately 20000 poses. After annotations, we were able to collect 10291 images with 3D pose annotations. Note that we remove `bad' poses and do not use them in the training. It took about 2-3 hours for an annotator to process 1000 images.

%% file: result.tex
\section{Experiments}

\subsection{Dataset and evaluation protocols}
\noindent {\bf MPII \cite{andriluka14cvpr} }
This dataset contains in-the-wild images for 2D human pose estimation, which includes 25k training images and 3k validation images. Those images are annotated with 2D joint locations and bounding boxes. In Section \ref{sec:dataset}, we constructed a 3D pose dataset on top of MPII dataset by annotating 3D joints to images. This dataset is used in training.% We tested our technique with 3k validation dataset and evaluated with the measure called PCK@0.5 \cite{andriluka14cvpr}.  

\noindent {\bf Human3.6M }
Human 3.6M dataset \cite{ionescu2014human3} is used in training and testing. Human 3.6M dataset is a large scale dataset for 3D human pose detection. This dataset contains 3.6 million images of 15 everyday activities, such as walking, sitting and making a phone call, which is performed by 7 professional actors. 3D positions of joint locations captured by motion capture (Mocap) systems are also available in the dataset. In addition, 2D projections of those 3D joint locations into images are available. The images are taken from four different views. As with previous researches, we down-sampled the video from 50fps to 10fps in order to reduce redundancy in video frames. We followed the same evaluation protocol used in previous approaches \cite{PavlakosZDD16,zhou2017weakly} for evaluation, where we use 5 subjects (S1, S5, S6, S7, S8) for training and the rest 2 subjects (S9, S11) for testing.  We used the 3D model of an actor provided in Human3.6 to generate mesh animations but this could be replaced with any 3D models.

The error metric used is called mean per joint position error (MPJPE) in $mm$. In the evaluation protocol, the position of the root joints is aligned with the ground truth but the global orientation is kept as is. Following \cite{zhou2017weakly} the output joint positions from ConvNets is scaled so that the sum of all 3D bone lengths is equal to that of a canonical average skeleton. This is done by:
\begin{equation}
{\bf P}_j = ({\bf P}_j^{\rm pred} - {\bf P}_{0}^{\rm pred}) \cdot l^{\rm ave} / l^{\rm pred} 
\end{equation}
where ${\bf P}_j^{\rm pred}$ is the predicted position, ${\bf P}_{0}$ is the root position, $l^{\rm pred}$ is the sum of skeleton length of the predicted skeleton and $l^{\rm ave}$ is the average of sum of skeleton length for all the training subjects in Human 3.6M dataset. 

We also evaluated the method with the error measure called the reconstruction error, where, before calculating the error, the result is aligned to the ground truth with a similarity transformation. 

%\noindent {\bf MPI-INF-3DHP }
%MPI-INF-3DHP dataset \cite{mehta2017vnect} contains images and 3D joints captured using MoCap. The images were captured by a MoCap system both in indoor and outdoor scenes. We only use its test set for evaluation, which contains 2929 frames from 6 subjects. 

%Following [15], we employ average PCK (with a threshold 150mm) and AUC as the evaluation metrics, i.e., after aligning the root joint (pelvis). Note that we assume the global scale is known for experimental evaluation. We observe that the definition of pelvis position in MPI-INF-3DHP is different from the one used in our training sets (i.e., Human 3.6M and MPII), so we moved the pelvis and hips towards neck
\subsection{Baselines}
Four baseline methods are implemented to conduct ablation studies. We trained the first three networks using Human3.6M dataset only and the fourth one with both Human3.6M and MPII dataset. 

\noindent {\bf Rotation regress (Rot reg) only } This only uses bone rotation regressor. The position is computed from linear integration of bone rotations. 

\noindent {\bf Heatmap (HM) only } This on the other hand only uses cross heatmap regressor.  

\noindent {\bf Rot reg + HM } This method is our proposed method which combines bone rotation regressor and cross heatmap regressor.

\noindent {\bf All } This is our proposed method trained using Human 3.6M and MPII dataset with 3D annotations obtained using the method presented in Sec \ref{sec:dataset}.

\subsection{Implementation and training detail}
Our method is implemented using MatConvNet toolbox \cite{vedaldi15matconvnet}. For the bone rotation regressor, we use ResNet50 \cite{DBLP:journals/corr/HeZRS15} as the base network, which is pre-trained on the ImageNet dataset. A single up-sampling layer and a single hourglass module is followed by the base network to predict heatmaps. We also use skip connections to connect skeleton regression layers and up-sampling layers. We used a skeleton with 16 joints and 15 segments each of them have 9 rotational parameters. The definition of joints is same as that of MPII dataset. Training a whole model takes about 1 day using three NVIDIA Quadro P6000 graphics cards with 24 GB memory. The batch size is 30 for each GPU. As for augmentation, we used left/right flip only---no scaling and rotation augmentation is used. We trained a model with SGD for 70 epochs, starting from the learning late of 0.001 and decreasing it to 0.00001. During test time, a single forward pass of our network is approx. 0.12 sec, which means the performance of our method is approx. 8-9 fps.    

We set the parameters in the loss function as $\alpha = \beta = \gamma = 0.1$ and $\lambda= 0.001$. When training with both Human 3.6M and MPII datasets, we randomly selected the images from both dataset such that half of the batch is filled with Human 3.6M and the other half by MPII dataset, following \cite{zhou2017weakly}. Since the 3D annotations of MPII are not accurate, we use them for supervising bone rotation regressor only. Thus, when training images are from MPII, we do not back propagate gradients from the losses that include the final 3D pose, ${\bf h}^{zy},$ ${\bf x}$ and ${\bf R}$, (i.e., $L_{pos}$, $L_{Rot}$ and the right term of $L_{hm}$) to the network. On the other hand, when the training sample is from Human3.6M dataset, which has accurate 3D annotations, we minimize all the losses.

\begin{figure}[tb]
	\centering
	\includegraphics[width=1\linewidth]{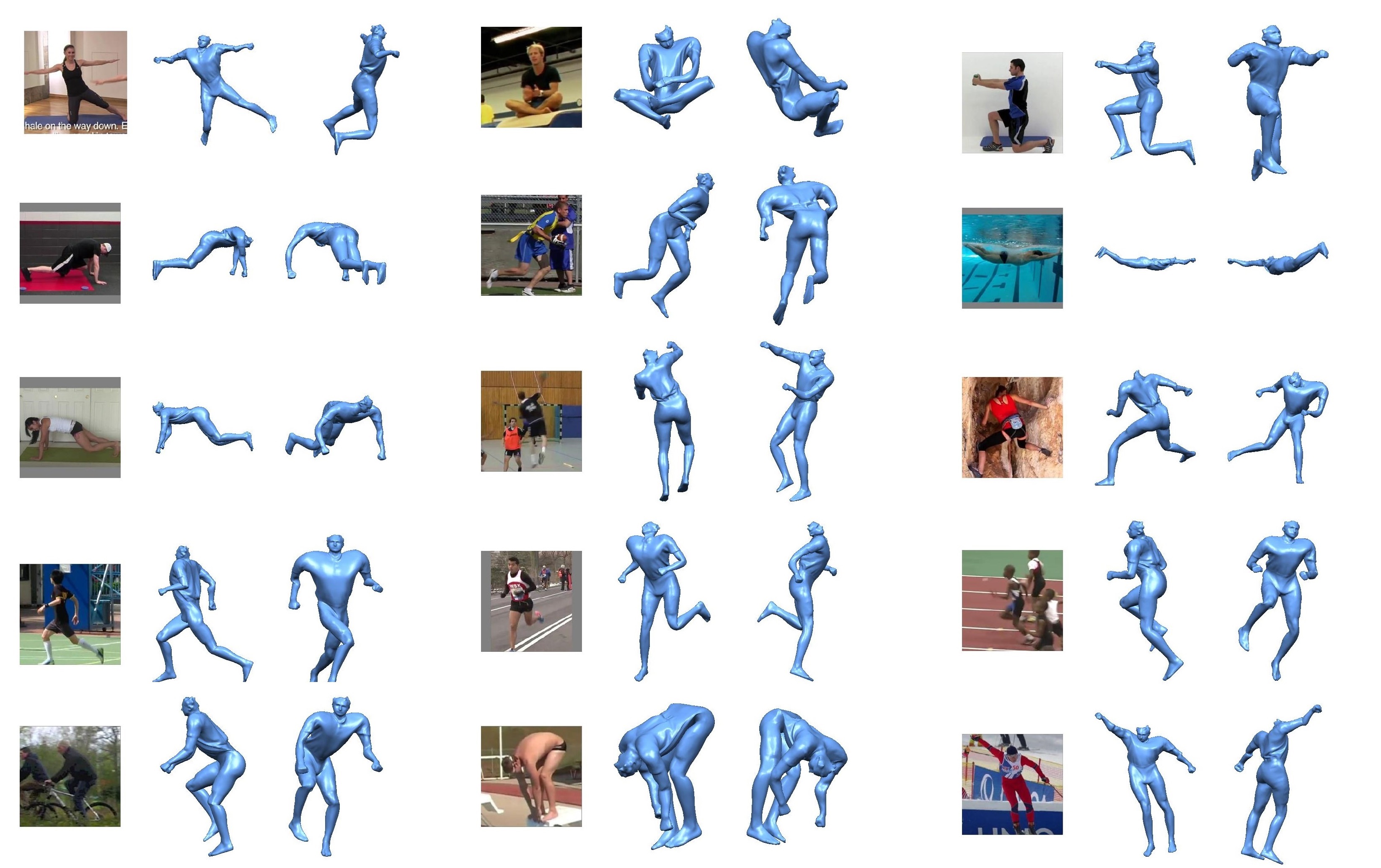}
	\caption{Some results on in-the-wild images. }
	\label{fig:result_mesh}
\end{figure}

\subsection{Results}
\subsubsection{Qualitative results}

In Figs. \ref{fig:result_mesh} and \ref{fig:result_mesh2}, we show the result of our 3D pose prediction method. Our method is able to predict 3D joint positions and bone orientations reasonably accurately even on in-the wild-images. Because our method can predict bone rotations of human body skeletons, we can produce mesh animations from 3D joint positions and rotations using linear blend skinning. Note that the rotations of hands and feet are not supervised.

\subsubsection{Comparisons to other state-of-the-art}

We have compared our techniques against other state-of-the-art in Table \ref{tab:comparison1}. Our technique is comparable with other state-of-the-art in terms of MPJPE.  Volumetric heatamaps \cite{PavlakosZDD16} can achieve MPJPE approx. 71 mm. However, it got worse results when including MPII (MPJPE 78 mm) with their decoupled structure, whereas we are around MPJPE 70 mm. Also, compared with \cite{PavlakosZDD16} with two stacks of hourglass networks, cross heatmap is more compact, which requires 1/32 of memory spaces to store. 

Compared with the previous techniques that predict angular poses \cite{zhou2016deep,hmrKanazawa17}, SkeletonNet is more accurate. In fact, MPJPE of our result is $69.9~{\rm mm}$, whereas that of Kanazawa et al. \cite{hmrKanazawa17} is $87.97~{\rm mm}$. In Table \ref{tab:recon_err}, we also compared the reconstruction error with previous approaches. Our technique outperforms previous techniques that iteratively optimizes joint angles \cite{bogo2016keep} and perform regression of joint angles \cite{pavlakos2018humanshape}. The benefit of SkeletonNet is, in addition to estimating 3D joint positions relatively accurately, we can predict 3D bone rotations, which is useful in animating a human body mesh or possibly predicting dynamics such as joint torques.

\begin{table}[hbt]
\footnotesize
\begin{center}
\caption{Comparisons to other state-of-the-art. MPJPE [mm] is used for error metric. }
\label{tab:comparison1}
\scalebox{0.85}{
\begin{tabular}{c c c c c c c c c}
\hline & Directions & Discussion & Eating & Greeting & Phoning & Photo & Posing & Purchases \\ 
\hline Zhou et al. \cite{zhou2016deep}    & 91.8        &  102.4      & 97.0        &   98.8      & 113.4     & 90.0        &  93.8    & 132.2  \\  
Tome et al. \cite{tome2017lifting} & 64.98 & 73.47 &  76.82 & 86.43 & 86.28 & 110.67  & 68.93 & 74.79 \\    
Mehta et al. \cite{mehta2016monocular} &  59.69  & 69.74  & 60.55 & 68.77 &  76.36 & 85.42 & 59.05 &  75.04\\      
Pavlakos et al. \cite{PavlakosZDD16} &  67.38 & 71.95 & 66.70 &   69.07 & 71.95 &  76.97 & 65.03 & 68.30 \\   
Ours (All)  & 63.33  & 71.59  & 61.39 & 70.40 & 69.90 &  83.17 & 62.98  & 68.77    \\  

% Zhou et al. & 87.36 &  109.31 &   87.05 &  103.16 & 116.18 &  143.32 &  106.88 &  99.78 \\                    
 
\hline & Sitting & SittingDown & Smoking & Waiting & WalkDog & Walking & WalkPair & Average \\ 
\hline Zhou et al. \cite{zhou2016deep}   &  159.0        &  106.9 & 125.2 & 94.4 & 79.0 & 126.0  & 99.0  & 107.3 \\  
Tome et al. \cite{tome2017lifting} & 110.19  & 172.91  & 84.95 & 85.78 & 86.26 & 71.36 & 73.14 & 88.39 \\      
Mehta et al. \cite{mehta2016monocular} & 96.19 & 122.92 &  70.82 &  68.45 & 54.41 & 82.03  & 59.79 & 74.14\\       
Pavlakos et al. \cite{PavlakosZDD16} & 83.66 & 96.51 &  71.74 & 65.83  & 74.89 & 59.11 & 63.24  & 71.90 \\        
Ours (All)  & 76.81 & 98.90 & 68.24 & 67.45 & 73.74 & 57.72 & 57.13 & 69.95  \\  
\hline 
\end{tabular}
}
\end{center}
\end{table}

\begin{table}[hbt]
	\footnotesize
	\begin{center}
		\caption{Comparison of reconstruction errors on Human 3.6M dataset. }
		\label{tab:recon_err}
		\scalebox{0.85}{
			\begin{tabular}{c c c c c c c c c}
				\hline Zhou et al. \cite{zhou2016sparseness} ~&~  Bogo et al. \cite{bogo2016keep} ~&~ Lassener et al. \cite{Lassner:UP:2017}~ &~ Pavlakos et al. \cite{pavlakos2018humanshape}~ & ~Ours \\ 
				\hline 
				106.7 & 82.3 & 80.7 & 75.9 & 61.4 \\
				\hline   
			\end{tabular}
		}
	\end{center}
\end{table}

\begin{table}[hbt]
	\begin{center}
		\caption{Comparisons between baselines. MPJPE [mm] is used for error metric. }
		\label{tab:comparison2}
		
		\begin{tabular}{c c c c }
			\hline 
			Rot reg only \;\;\;& Heatmap only \;\;\; &	Rot reg + HM \;\;\; & All \\ 
			\hline  112.43 \;\;\;   & 128.55 \;\;\;  & 87.05 \;\;\; & 69.95 \\ 
			\hline 
		\end{tabular}
		
	\end{center}
\end{table}

\subsubsection{Comparisons between baselines}

\begin{figure}[tb]
	\centering
	\includegraphics[width=1\linewidth]{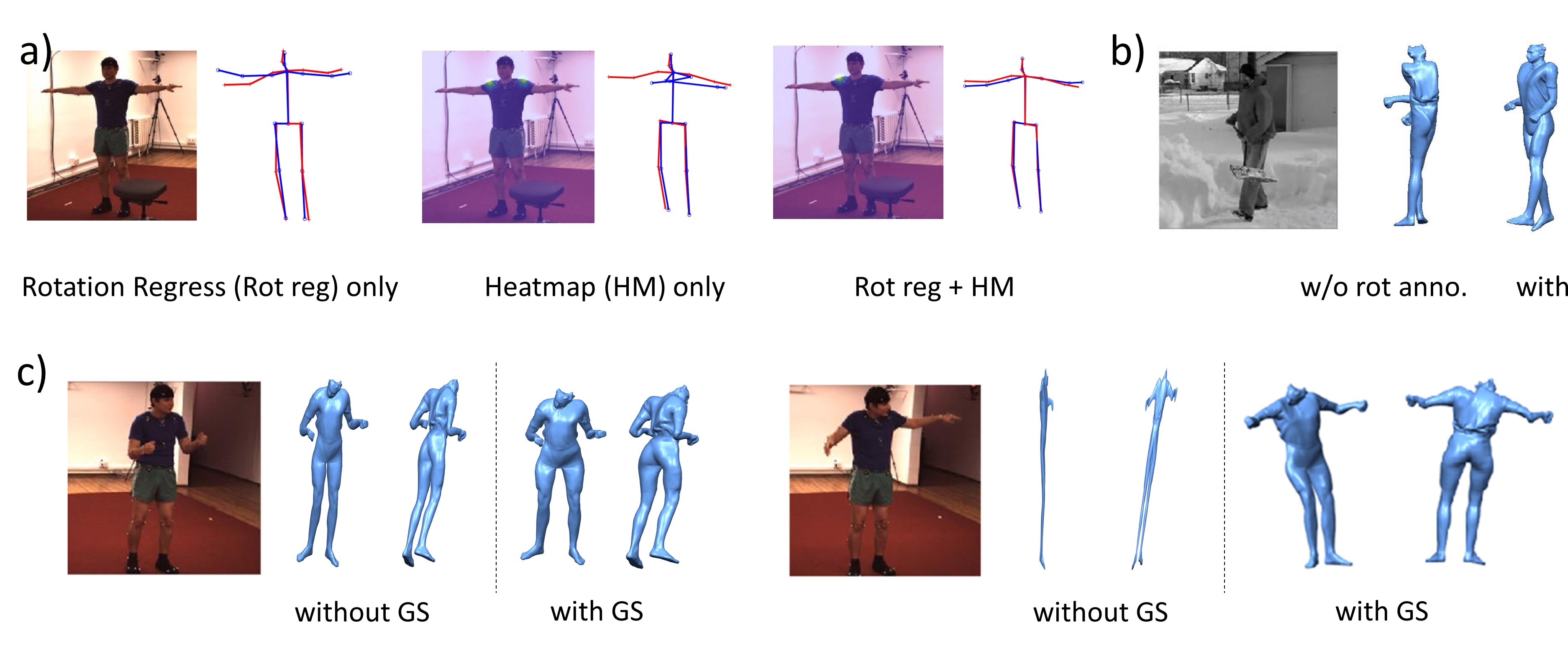}
	\caption{Comparisons of baselines. a) Bone rotation regressor preserves skeletal structure but the joint positions are not accurate enough. With only heatmaps, skeletal structure is sometimes destructed e.g., with the left and right flips. By combining both, it produces more accurate result while preserving structure. b) The rotation annotation is important for reconstructing a pose from in the wild images. c) With the proposed Gram Schmidt (GS) orthogonalization layer, undesirable deformations such as shears and scalings are removed. }
	\label{fig:Ablation_study}
\end{figure}

In  Table \ref{tab:comparison2} and Fig. \ref{fig:Ablation_study}, we show comparisons between the baselines. As can be seen from Fig. \ref{fig:Ablation_study} a, the result of bone rotation regressor preserves skeletal structure, but the joint positions are not accurate enough. With only heatmaps, however, skeletal structure is sometimes destructed e.g., by the left and right flips. By combining our bone rotation and cross heatmap regressor, a more accurate result is produced, while preserving skeletal structure. Note that in previous work this kind of confusions are remedied by incorporating recurrence \cite{belagiannis2016recurrent} or using many stacks of a hourglass module \cite{newell2016stacked}. By training with both Human 3.6M and MPII, we get the best result (Table \ref{tab:comparison2}). In addition, we found that annotating 3D rotations is important for reconstructing human poses from in-the-wild images (Fig. \ref{fig:Ablation_study} b). Thus, the key to our improvements in MPJPE is the use of cross heatmap and the use of MPII dataset in training. Even when MPII dataset is not provided for training, SkeletonNet can predict reasonably accurately 3D human pose by exploiting the combination of skeletal structure and heatmaps.

%Zhou et al. \cite{zhou201} & 124.52 & 199.23 & 107.42  &  118.09 & 114.23  & 79.39 & 97.70  & 79.9 \\        

\subsubsection{Comparisons between rotation representation}
We have also compared the results of bone rotation regressor by changing its rotation representation. Specifically, we tested the network that 1) regressess axis angles but indirectly supervised with rotation matrices (AA), 2) regressess axis angles but supervised with relative joint rotation matrices and converts them back to the absolute space using forward kinematics, which is equivalent as SMPL \cite{bogo2016keep}  (FKAA), 3) regresses absolute rotations (AbsRotReg), 4) regresses rotations without the GS layer (w/o GS), 5) regresses a global rotation (GlobalReg), 6) classifies a global rotation (GlobalClass) and 7) is same as GlobalClass but aligns rotations with heatmaps (All), respectively. The networks are trained with Human 3.6M, except for All that was trained with both MPII and Human 3.6M. To compare the rotation prediction accuracy, we compute relative rotations between the ground truth and predicted bone rotations, convert them to axis angles and take the norms in degrees, which reflects all three DoFs of rotations.

In Table  \ref{tab:cmp_rotation}, Global Rot. Err. indicates the error of global rotations. Bone Rot. Err. indicates the average error of bone rotations relative to the root. As shown in Table \ref{tab:cmp_rotation}, the proposed method based on the GS layer, which  classifies a global rotation, is the best in terms of MPJPE accuracy. AbsRotReg is also high in accuracy but it produces bone rotations with its determinant of -1, which collapse skeletal structure. The method based on axis angle tends to produce large errors probably because of their high non-linearity, requiring an iterative process \cite{hmrKanazawa17} or a more informative geometric loss, e.g., the one using differences between silhouettes \cite{pavlakos2018humanshape}. In summary, our method can benefit from the use of $3 \times3$ rotation matrices, which can probably be modeled as simpler functions than other angle representations, which is more friendly to ConvNets to learn with. As reported in \cite{zhou2016deep}, we could not train a network properly using Euler angles as supervisions, where training and validation losses remained high. In contrast, SkeletonNet can model subtle pose appearances due to e.g., medial and lateral rotations around segment axes by providing supervisions on both rotations and positions. With joint position supervisions only and no rotational supervisions, it is possible to get reasonable results in joint position predictions \cite{zhou2016deep} but is difficult to obtain good results for bone orientations.

\begin{table}[hbt]
	\footnotesize
	\begin{center}
		\caption{Comparison of rotation representation. }
		\label{tab:cmp_rotation}
		\scalebox{0.8}{
			\begin{tabular}{c c c c c c c c c}
				\hline ~&~ ~&~AA ~&~ FKAA ~&~AbsRotReg ~&~ w/o GS ~&~ GlobalReg ~ & ~ GlobalClass ~ & ~ All \\ 
				\hline 
				& MPJPE            & 175.06 & 197.44 & 114.70 & 124.06 & 119.18 & 112.42 & 69.95 \\
				& Global Rot. Err. &  30.46 & 35.81  & 18.83  & 21.28  & 21.64  & 20.81  & 12.93 \\
				& Bone Rot. Err.   & 37.34 & 44.77  & ---    & 28.72  & 29.08  & 29.46  & 21.94 \\
				\hline   
			\end{tabular}
		}
	\end{center}
\end{table}

\begin{comment}
\begin{table}[hbt]
	\footnotesize
	\begin{center}
		\caption{Results on MPII validation dataset. PCK@0.5 is used for the error measure. }
		\label{tab:PCK}
		\scalebox{0.9}{
			\begin{tabular}{c c c c c c c c c}
				\hline & Head & Sho. & Elbow & Wrist & Hip & Knee & Ankle & Mean \\ 
				\hline 
				Pavlakos et al. (4 stacks) \cite{PavlakosZDD16} &  95.86& 93.05  &	85.73  & 80.42 & 81.29 & 79.31 & 75.41 & 84.96 \\   
				Ours ( $\approx$2 stacks)  & 93.86 & 89.2 & 71.18  & 79.92  & 74.49  & 72.15 & 79.85 & 79.99 \\
				\hline   
			\end{tabular}
		}
	\end{center}
\end{table}

\end{comment}

%57.22 69.59 58.84 64.76 66.42 76.21 57.31 63.54 81.72 118.81  66.50 64.85 73.20 54.50 52.87 68.42
%88.1651   83.6277   75.6435   70.5188   69.0671   72.3554   71.0430   76.5967   76.0786

\subsubsection{Failure cases and limitations}
\begin{figure}[tb]
	\centering
	\includegraphics[width=1\linewidth]{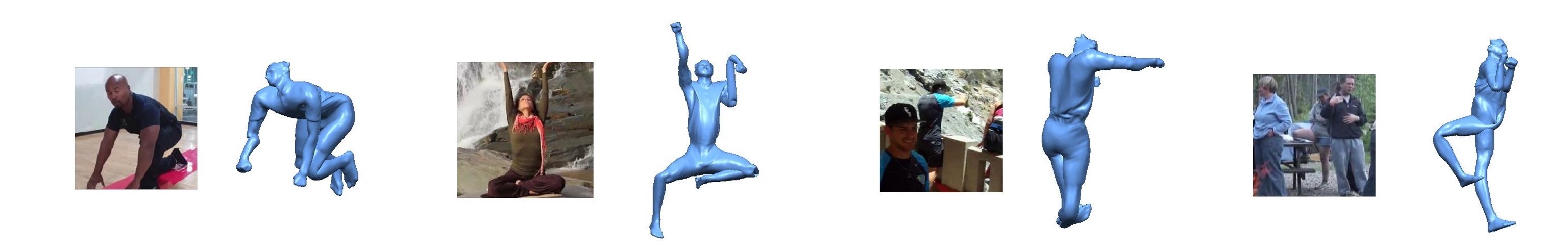}
	\caption{Failure cases.}
	\label{fig:failure}
\end{figure}

In Fig. \ref{fig:failure}, we show failure cases. Our technique fails when there are large self-occlusions and occlusions by objects or other humans. In addition, our network are currently designed for the single-person detection and thus fails when multiple humans exist in the image. Since we scale a skeleton, we are not be able to model absolute bone lengths. Cross heatmap regressor possesses the ability to alter relative bone lengths but our method have generalization issues when the body type is extremely different from the original skeleton, e.g., prediction of small children's poses. Also our network does not take into account hand and foot orientations.

%If the bone rotation regressor produces good enough results for the initial solution to cross heatmap regressor, 

%% file: conclusion.tex
\section{Conclusion}

We have presented SkeletonNet, a novel end-to-end 3D human pose detection technique from a single image. The first step regresses bone segment rotations to obtain an initial solutions without large errors by considering skeleton structure. The second step performs refinement based on heatmap regressor that is based on the representation called cross heatmap which stacks heatmaps of xy and zy coordinates. This combination allows us to predict bone orientations and joint positions accurately, which may provide useful information to applications like animation and biomechanics. We also presented a 3D human pose dataset constructed by adding 3D rotational annotations to publicly-available 2D human pose dataset.

In future work, we would like to address monocular detections of other human body properties, such as body shape, body weight, contact forces and joint forces/torques. We are also interested in generative adversarial networks (GAN) to improve pose prediction results using an unsupervised  manner based on the image dataset that does not have 3D annotations.